\documentclass[10pt,twocolumn,letterpaper]{article}

\usepackage{wacv}              

\usepackage{multirow}
\usepackage{array}
\usepackage{subcaption}
\usepackage{amsmath,amsfonts}

\usepackage{xcolor}
\definecolor{RoyalBlue}{RGB}{65, 105, 225}
\definecolor{TiffanyBlue}{RGB}{10, 186, 181}

\definecolor{wacvblue}{rgb}{0.21,0.49,0.74}
\usepackage[pagebackref,breaklinks,colorlinks,allcolors=wacvblue]{hyperref}

\def\wacvPaperID{*****} 
\def\confName{WACV}
\def\confYear{2026}

\title{MedProbCLIP: Probabilistic Adaptation of Vision-Language Foundation Model for Reliable Radiograph-Report Retrieval}

\author{\normalsize
\textbf{Ahmad Elallaf}$^1$\hspace{1.25mm}
\textbf{Yu Zhang}$^2$\hspace{1.25mm}
\textbf{Yuktha Priya Masupalli}$^1$\hspace{1.25mm} 
\textbf{Jeong Yang}$^1$\hspace{1.25mm}
\textbf{Young Lee}$^1$\hspace{1.25mm}
\textbf{Zechun Cao}$^1$\hspace{1.25mm}
\textbf{Gongbo Liang}$^{1*}$\vspace{.75mm}\\
{\normalsize$^1$ Texas A\&M University-San Antonio~~~
$^2$ Boise State University
}\\
{\normalsize 
aelallaf@tamusa.edu~~~~~
yzhang@boisestate.edu~~~~~
ymasupalli@tamusa.edu~~~~~
\{jyang, ylee, zcao, gliang\}@tamusa.edu}\vspace{.75mm}\\
{\tt \normalsize \url{https://github.com/FOURM-LAB/MedProbCLIP}}
}

\begin{document}
\maketitle
\begin{abstract}
Vision-language foundation models have emerged as powerful general-purpose representation learners with strong potential for multimodal understanding, but their deterministic embeddings often fail to provide the reliability required for high-stakes biomedical applications. This work introduces MedProbCLIP, a probabilistic vision-language learning framework for chest X-ray and radiology report representation learning and bidirectional retrieval. MedProbCLIP models image and text representations as Gaussian embeddings through a probabilistic contrastive objective that explicitly captures uncertainty and many-to-many correspondences between radiographs and clinical narratives. A variational information bottleneck mitigates overconfident predictions, while MedProbCLIP employs multi-view radiograph encoding and multi-section report encoding during training to provide fine-grained supervision for clinically aligned correspondence, yet requires only a single radiograph and a single report at inference. Evaluated on the MIMIC-CXR dataset, MedProbCLIP outperforms deterministic and probabilistic baselines, including CLIP, CXR-CLIP, and PCME++, in both retrieval and zero-shot classification. Beyond accuracy, MedProbCLIP demonstrates superior calibration, risk-coverage behavior, selective retrieval reliability, and robustness to clinically relevant corruptions, underscoring the value of probabilistic vision-language modeling for improving the trustworthiness and safety of radiology image-text retrieval systems.

\end{abstract}
    
\section{Introduction}
Modern neural network models have shown promising results in various application domains, particularly in healthcare~\cite{liu2022llrhnet,liang2022development,liu2023simulated} and safety-critical systems~\cite{liang2023unveiling,zulu2024enhancing,elallaf2026betarisk,lee2025machine}. However, the success of these models is often constrained by their dependence on large amounts of fully labeled training data.

To mitigate this data scarcity, vision-language representation learning has emerged as a powerful paradigm. By modeling the semantic relationships between visual observations and their associated textual descriptions~\cite{radford2021learning}, these frameworks learn robust, transferable embeddings that support diverse downstream applications~\cite{endo2021retrieval,wang2022medclip,you2023cxr} without requiring dense manual annotations. In the biomedical domain, aligning chest X-rays with radiology reports creates a unified multimodal representation of clinical information~\cite{zhang2023knowledge}. Cross-modal retrieval is particularly impactful in this context.

Despite this potential, medical image-text retrieval poses unique challenges compared to natural image domains. Radiographs frequently exhibit subtle findings, multi-view inconsistencies, and overlapping disease patterns, while free-text reports describe these observations with varying levels of specificity. Consequently, the correspondence between images and reports is inherently \textit{many-to-many}: a single report may summarize findings spanning multiple studies, and the same pathology can manifest across distinct radiographs. Conventional contrastive frameworks, such as CLIP~\cite{radford2021learning}, TIMNet~\cite{liang2021contrastive}, and CXR-CLIP~\cite{you2023cxr}, treat alignment as deterministic, point-to-point matching, limiting their ability to capture this structured ambiguity.

Beyond structural ambiguity, reliability is crucial for clinical adoption~\cite{jiang2011calibrating}. Deterministic embeddings produce overconfident similarity scores and cannot express when an image-report alignment is uncertain~\cite{guo2017calibration,liang2020imporved}. Yet trustworthy medical AI systems must offer calibrated confidence estimates, support selective prediction (i.e., abstaining on uncertain cases), and remain robust to variations in image quality, patient positioning, and acquisition conditions. Existing models rarely incorporate explicit uncertainty modeling and therefore risk generating brittle or misleading retrieval results~\cite{liang2025uncertainty}.

To address these limitations, we introduce \textbf{MedProbCLIP}, a probabilistic vision-language framework for chest X-ray and radiology report representation learning. Rather than mapping inputs to single embedding points, MedProbCLIP learns distributional representations, estimating both a mean and variance for each image and text input. This probabilistic formulation naturally captures predictive uncertainty and more faithfully reflects the complex, many-to-many relationships present in clinical data.

To the best of our knowledge, this work presents one of the first systematic studies in medical vision-language retrieval demonstrating that probabilistic modeling simultaneously improves retrieval accuracy and reliability. Our contributions are summarized as follows:

\begin{itemize}
    \item We introduce MedProbCLIP, a probabilistic contrastive learning framework for medical image-text retrieval that models similarities using distribution-based embeddings rather than deterministic points.

    \vspace{.75mm}
    \item We perform a comprehensive evaluation against strong and widely adopted baselines, including CLIP, CXR-CLIP, and PCME++~\cite{chun2023pcmepp}, under identical training and evaluation conditions on the MIMIC-CXR dataset~\cite{johnson2019mimic}.

    \vspace{.75mm}
    \item We demonstrate that probabilistic modeling consistently improves retrieval performance while enhancing uncertainty-aware reliability, including better calibration and selective prediction capabilities compared to deterministic counterparts.
\end{itemize}

\section{Background}
\vspace{-1.mm}
\begin{figure}
    \centering
    \includegraphics[width=0.975\linewidth]{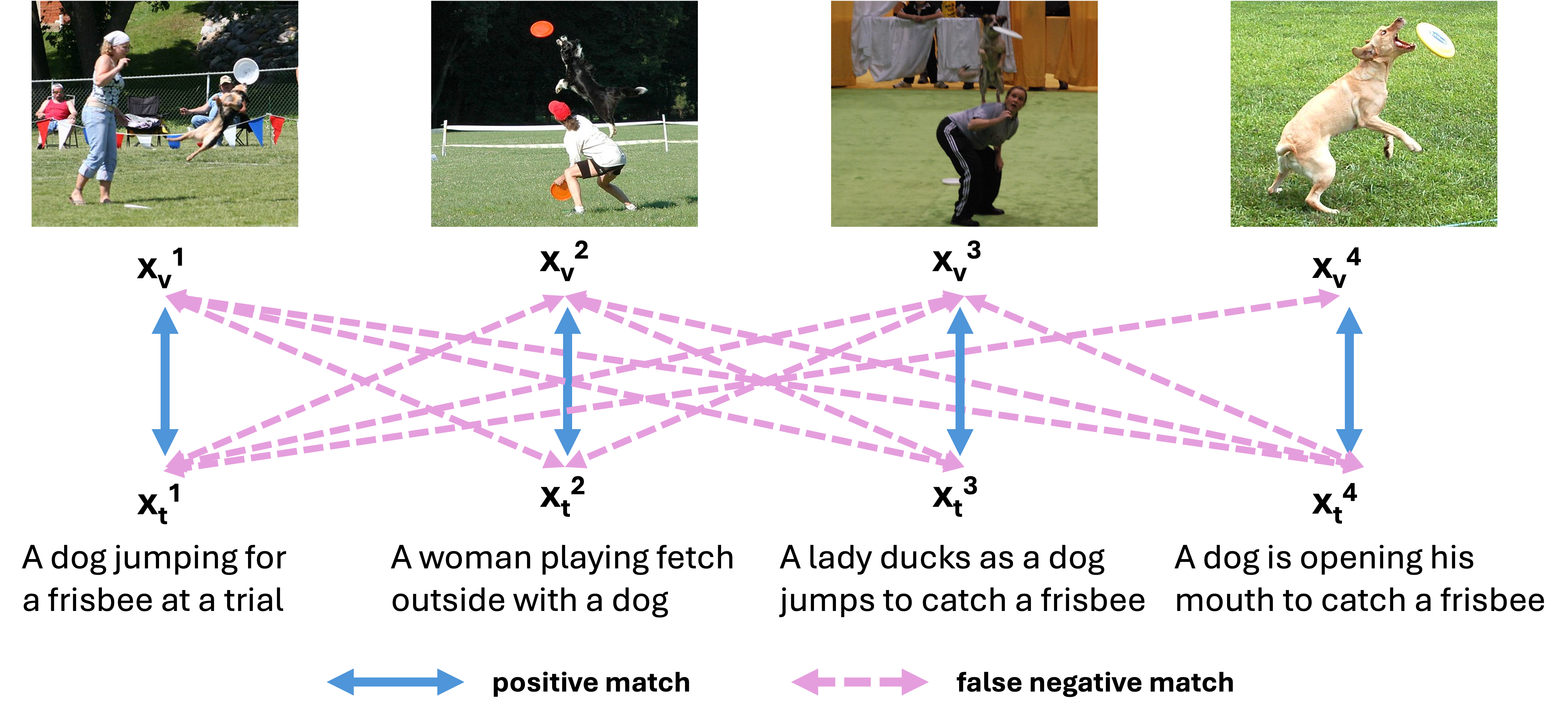}
    \vspace{-1.5mm}
    \caption{Illustration of inherent many-to-many relationships in cross-modal datasets. Although MS-COCO annotates only a single caption as the positive match to one image (blue arrows), human raters often identify multiple additional plausible matches (pink dashed arrows). Such unannotated positives create false negatives that violate the one-to-one assumption commonly enforced in contrastive learning, motivating models capable of handling ambiguity and uncertainty in image-text alignment.}
    \vspace{-3.5mm}
    \label{fig:one-to-one_matchin}
\end{figure}

\subsection{Problem Definition}
\vspace{-.5mm}
Let a cross-modal retrieval dataset be defined as $X=\{(x_v^{(i)}, x_t^{(i)})\}^N_{i=1}$, where $x_v^{(i)}\in V$ and $x_t^{(i)}\in T$ denote paired samples from the visual and textual domains, respectively. Under the standard dataset construction, each pair $(x_v^{(i)}, x_t^{(i)})$ is treated as the only positive match, i.e., $y_{vt}^{(i, j)}=1 \iff i=j$, where $y_{vt}^{(i, j)}$ indicates whether the $i$-th image and the $j$-th text naturally correspond. Contrastive retrieval models are therefore trained to learn a matching function $h(\cdot)$ such that $h(x_v^{(i)}, x_t^{(i)}) > x_v^{(i)}, x_t^{(j)}, \forall i\neq j$, implicitly enforcing a one-to-one alignment between modalities.

However, real data rarely obey this assumption.
Figure~\ref{fig:one-to-one_matchin} illustrates that many datasets, such as MS-COCO, contain numerous plausible yet unannotated positive matches. Humans often judge multiple $x_t^{(j)}$ as relevant to the same $x_v^{(i)}$, and vice versa, resulting in inherently many-to-many relationships. In medical imaging, this issue becomes more pronounced: a radiology report may summarize findings across several studies, and a single pathology can appear across multiple images from the same or different encounters. Thus, datasets built with strict one-to-one assumptions inevitably contain large numbers of false negatives, which can misguide deterministic contrastive learning approaches.

\subsection{Contrastive Learning}
\vspace{-.5mm}


\begin{figure}[!tb]
\centering
\includegraphics[width=0.475\textwidth]{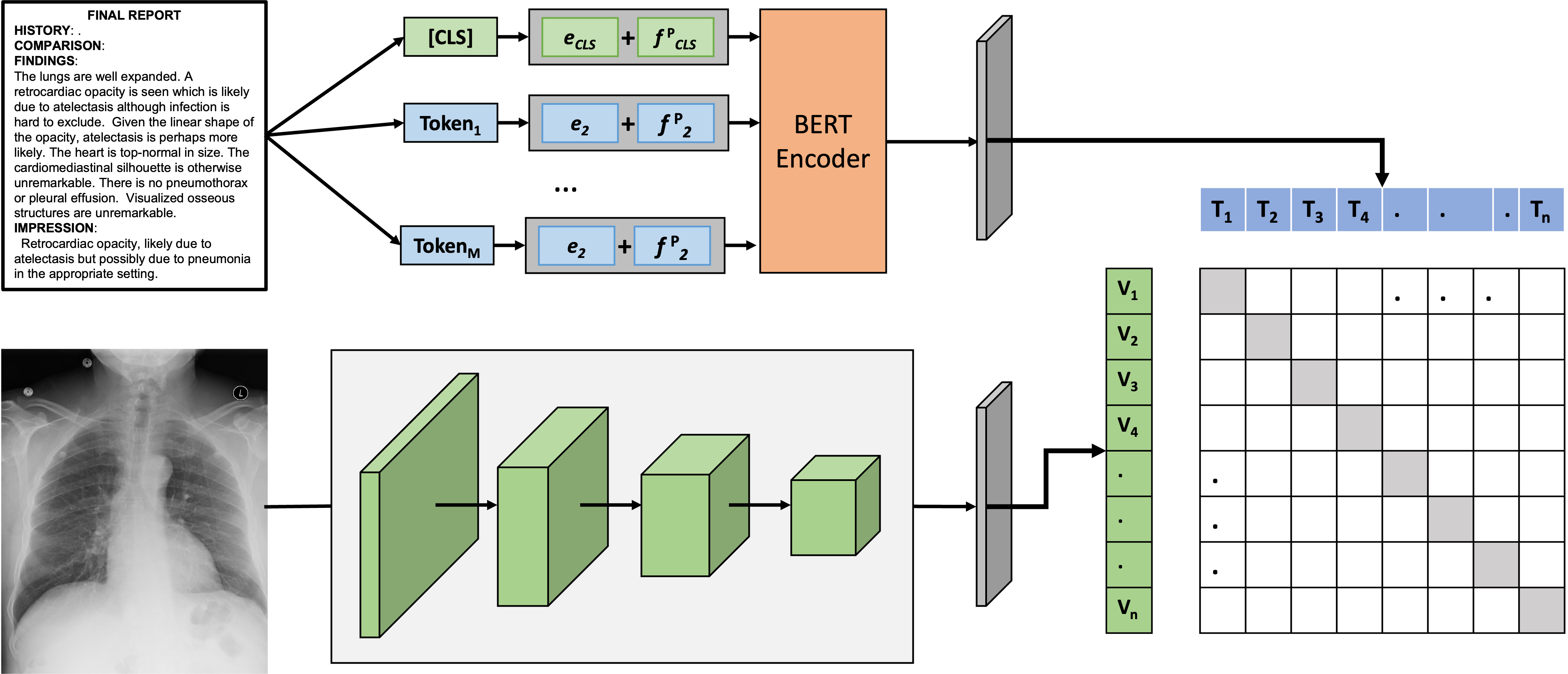}
\caption{Example of a contrastive learning model.}
\vspace{-3.5mm}
\label{fig:contrastive_learning}
\end{figure}

Contrastive learning provides a principled framework for learning joint representations across modalities.
Given an image-text pair $(x_v^{(i)}, x_t^{(i)})$, dual-encoder models learn embeddings $z_v^{(i)}=h_v(x_v^{(i)})$ and $z_t^{(i)}=h_t(x_t^{(i)})$, where $h_v(\cdot)$ and $h_t(\cot)$ are modality-specific encoders. A similarity function $s(\cdot,\cdot)$ (e.g., cosine similarity) is used to encourage matched pairs to be close and mismatched pairs to be far apart: $s(x_v^{(i)}, x_t^{(i)}) \gg s((x_v^{(i)}, x_t^{(j)}), i\neq j$. This objective is typically implemented with a bidirectional InfoNCE loss or its variants, forming the basis of widely used models such as CLIP~\cite{radford2021learning}, ConVIRT~\cite{zhang2022contrastive}, and TIMNet~\cite{liang2021contrastive}.
Figure~\ref{fig:contrastive_learning} illustrates the general architecture: two encoders independently embed images and texts, and a contrastive loss aligns the paired representations.

While contrastive learning is highly effective, deterministic embeddings $z_v^{(i)}$ and $z_t^{(i)}$ assume a single correct alignment for each pair. This assumption breaks down under many-to-many relationships or ambiguous matches, where multiple texts may be relevant to the same image and vice versa (as shown in Figure~\ref{fig:one-to-one_matchin}). Under such circumstances, deterministic contrastive losses can be misled by false negatives, forcing embeddings of genuinely related pairs apart. This limitation motivates distributional or uncertainty-aware representations that can better capture the intrinsic ambiguity in real-world cross-modal data.


\subsection{Image-Text Retrieval in Medical Imaging}
\vspace{-.5mm}

Large-scale paired datasets such as MIMIC-CXR~\cite{johnson2019mimic} have enabled significant progress in cross-modal retrieval between chest X-rays and their associated radiology reports. 
A broad spectrum of medical vision-language models has been proposed, which can be roughly categorized by the supervision they require. Models with enhanced supervision, such as BioViL~\cite{bannur2023learning}, achieve strong performance by leveraging explicit phrase grounding: radiology reports are manually annotated with localized regions (e.g., bounding boxes) corresponding to specific clinical findings. While these methods often set the absolute state-of-the-art, their reliance on costly, fine-grained annotations places them in a fundamentally different supervision regime, as such annotations are rarely available in real-world clinical datasets.

In contrast, paired-only models learn directly from readily available image-report pairs without requiring additional structured labels. Approaches such as ConVIRT~\cite{zhang2022contrastive}, TIMNet~\cite{liang2021contrastive}, and CXR-CLIP~\cite{you2023cxr} follow this paradigm, with CXR-CLIP emerging as the strongest, most widely adopted representatives that follows the CLIP architecture, 
yielding robust retrieval performance while remaining fully compatible with the supervision level provided by MIMIC-CXR.

Because our work aims to improve models that learn solely from paired image-report data, without relying on phrase-grounding annotations or other enhanced supervision, CXR-CLIP represents the appropriate state-of-the-art baseline for our setting. It is strong, widely recognized, and directly comparable under identical training conditions.
\section{Method}
\begin{figure}[!tb]
    \centering
    \includegraphics[width=0.975\linewidth]{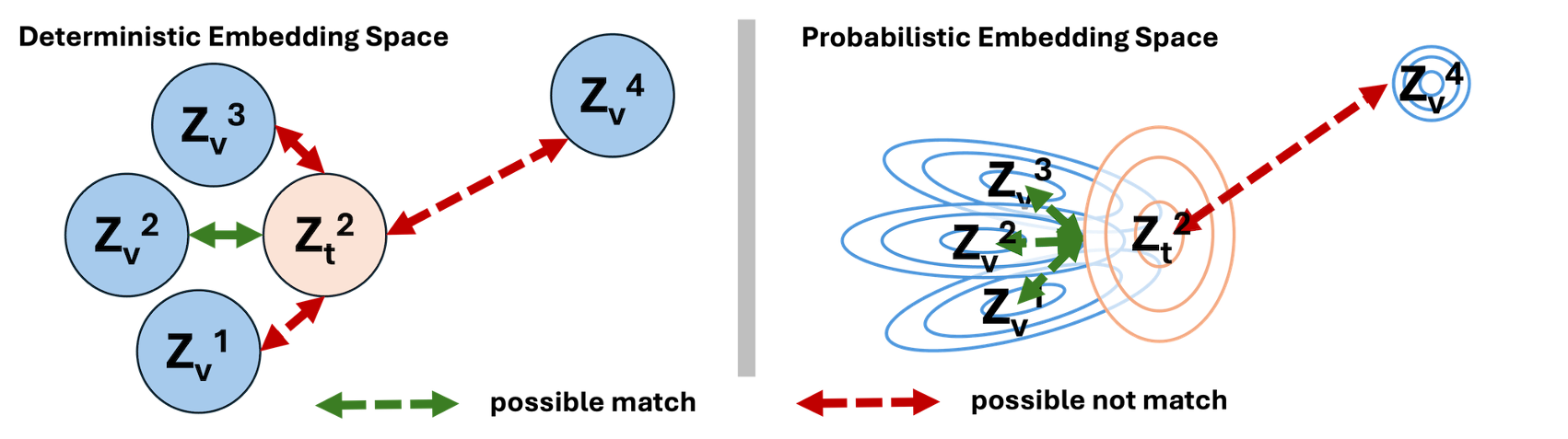}
    \vspace{-1.mm}
    \caption{Deterministic Embedding vs Probabilistic Embedding}
    \vspace{-2.5mm}
    \label{fig:determistic_vs_probabilistic}
\end{figure}
\vspace{-.5mm}
\subsection{Motivation}
\vspace{-.5mm}
Cross-domain retrieval models such as CLIP and CXR-CLIP learn a shared embedding space using deterministic point embeddings and contrastive losses. Although effective, these models map each input to a single location in latent space, which limits their ability to represent the ambiguity, annotation noise, and intra-class variability inherent in medical imaging.

Figure~\ref{fig:determistic_vs_probabilistic} (left) illustrates this limitation. Three visually similar chest X-rays ($Z_v^{(1)}$, $Z_v^{(2)}$, $Z_v^{(3)}$) lie close to the same text embedding ($Z_t^{(2)}$). A deterministic formulation forces the model to assign a single positive label $y_{vt}^{(2,2)}=1$ while treating other clinically plausible alternatives as negatives, despite their overlapping semantics. This produces noisy supervision, distorted gradients, and overconfident retrieval scores that accumulate severely in domains with subtle findings and many-to-many relationships.

\begin{figure}[!tb]
    \centering
    \includegraphics[width=0.975\linewidth]{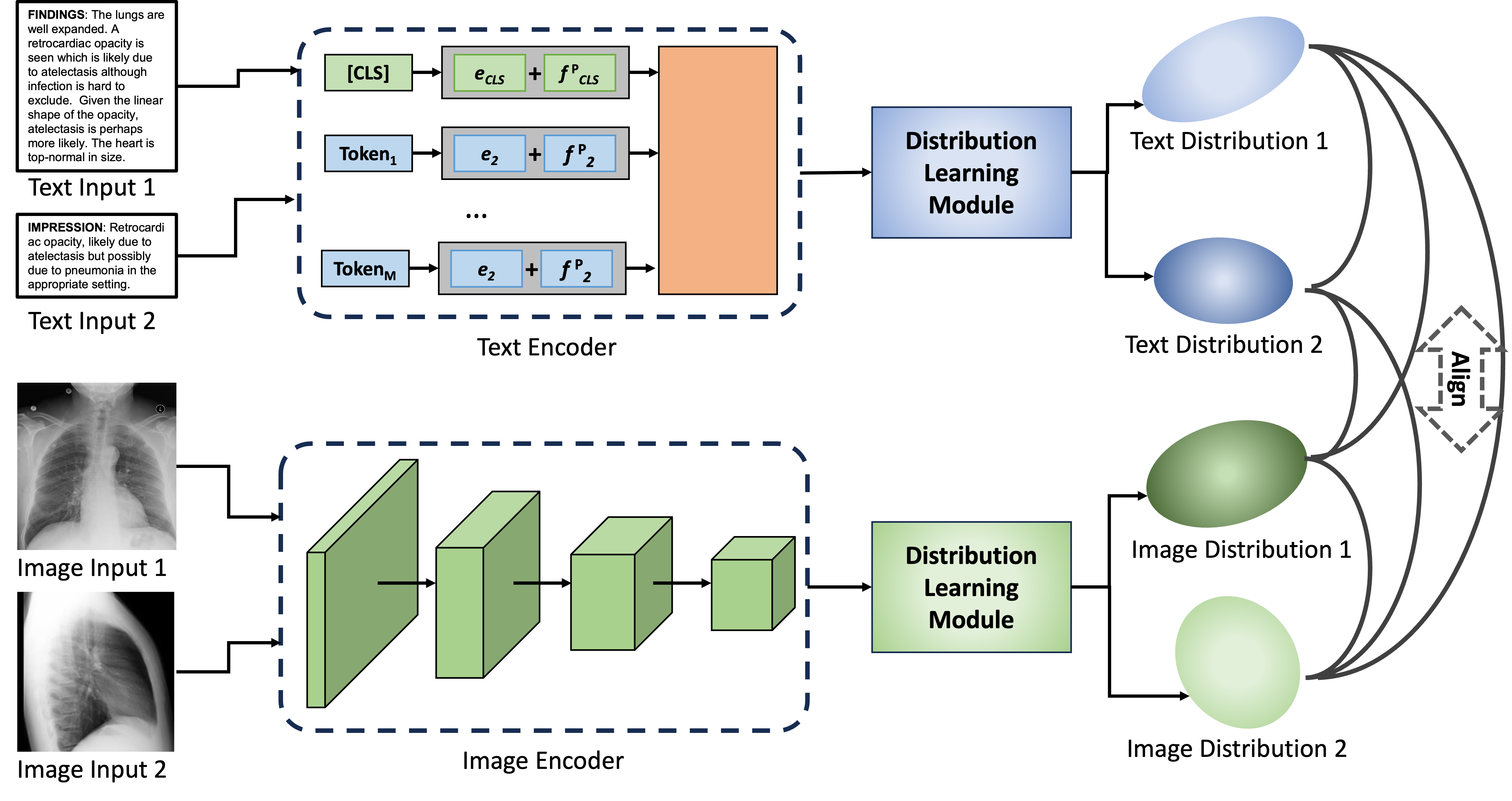}
    \vspace{-1.mm}
    \caption{Overview of the MedProbCLIP architecture. 
    }
    \vspace{-2.5mm}
    \label{fig:architecture}
\end{figure}

In contrast, probabilistic embedding represents each input as a distribution rather than a point. As shown in Figure~\ref{fig:determistic_vs_probabilistic} (right), ambiguous or semantically overlapping samples naturally exhibit larger variances, indicating uncertainty in their alignment. Conversely, confidently unmatched samples (e.g., $Z_v^{(4)}$) form compact, low-variance distributions.

\subsection{Probabilistic Contrastive Learning}
\vspace{-.5mm}
To capture uncertainty in cross-modal alignment, MedProbCLIP models image and text embeddings as diagonal Gaussian distributions. For an image $x_v$ and a report $x_t$, their latent representations are: $Z_v \sim \mathcal{N}(\mu_v, \sigma_v^2)$ and $Z_t \sim \mathcal{N}(\mu_t, \sigma_t^2)$, 
where $\mu_v, \mu_t \in \mathbb{R}^D$ are mean embeddings and $\sigma_v^2, \sigma_t^2 \in \mathbb{R}_{+}^D$ are diagonal variances. Low-variance distributions indicate confident matches, whereas high-variance embeddings represent ambiguity or weak evidence.



Following PCME++~\cite{chun2023pcmepp}, the discrepancy between two distributions is computed using the contrastive stochastic distance (CSD):
\begin{equation}
    \label{eq:csd}
    d(Z_v, Z_t) = 
    \frac{1}{2}
    \sum_{d=1}^{D}
    \left[
    \frac{\left(\mu_v^{(d)} - \mu_t^{(d)}\right)^2}{\sigma_v^{2(d)} + \sigma_t^{2(d)}}
    +
    \log\left(\sigma_v^{2(d)} + \sigma_t^{2(d)}\right)
    \right],
\end{equation}
which jointly accounts for both mean separation and uncertainty through the summed variances.

Given a binary match label $y_{vt} \in\{0,1\}$, the probabilistic negative log-likelihood (NLL) is:
\begin{equation}    
    \begin{aligned}
    \mathcal{L} = & -y_{vt} \log(\sigma(-a \cdot d(Z_v, Z_t) + b)) \\
             & - (1-y_{vt}) \log(\sigma(-a \cdot d(Z_v, Z_t) - b)),   
    \end{aligned}
    \label{eq:nll}
\end{equation}
where $\sigma(\cdot)$ denotes the sigmoid function, and $a,b$ are learnable scalars controlling the scale and offset of the logits. This objective encourages matched distributions to be close (and confident) while pushing unmatched ones apart.

To regularize variances and prevent trivial solutions, each distribution is constrained via KL divergence to a unit Gaussian:
\begin{equation}
    \label{eq:kl}
    \mathcal{L}_{\mathrm{KL}} = 
    \frac{1}{2}\sum_{d=1}^{D}
    \left(
    \sigma^{2(d)} + \left(\mu^{(d)}\right)^2 - 1 - \log\sigma^{2(d)}
    \right),
\end{equation}
computed separately for images and texts.


\subsection{Architecture}
 
Figure~\ref{fig:architecture} presents the overall architecture of our probabilistic contrastive learning framework for medical image-report retrieval. Unlike CLIP or PCME++, which operate on single image-text pairs, our model jointly processes two image inputs ($x_{v_1}, x_{v_2}$) and two text inputs ($x_{t_1}, x_{t_2}$). This design exploits two common properties of clinical data: 1) medical reports contain multiple semantically complementary sections (e.g., Findings, Impression), and 2) chest X-ray studies often include multiple imaging views (e.g., PA and lateral). When a second view or second text section is not available, we generate an auxiliary input through data augmentation to maintain consistent training behavior.

The model consists of two encoder branches: a visual encoder 
for chest X-ray images and a textual encoder 
for radiology reports. Each encoder produces a deterministic embedding, which is then transformed by a distribution learning module into a Gaussian distribution parameterized by mean $\mu$ and diagonal variance $\sigma^2$. This yields four probabilistic embeddings, $Z_{v_1}$, $Z_{v_2}$, $Z_{t_1}$, and $Z_{t_2}$, capturing both semantic content and uncertainty arising from ambiguity across views, report sections, or image-text relationships.

\subsection{Model Training}
During training, the model computes probabilistic distances between all relevant pairs of distributions, including image-text pairs, image-image pairs, and text-text pairs, using Equation~\ref{eq:nll} to compare Gaussian embeddings. 

To prevent the model from predicting unbounded variances and to regularize the distributional space, we additionally apply a variational information bottleneck (VIB) penalty (Equation~\ref{eq:kl}), computing KL divergence between each embedding distribution and a unit Gaussian prior.

The final training objective is a weighted combination of inter-modal NLL, intra-modal NLL terms, and KL regularization:
\begin{equation}
    \begin{aligned}            
    \label{eq:final_loss}
    \mathcal{L}_{\text{total}} = &
    \mathcal{L}_{\text{inter}} + \lambda_I \mathcal{L}_{\text{intra-I}} + \lambda_T \mathcal{L}_{\text{intra-T}} + \\
    & \beta_I\mathrm{KL}_{\text{img}} + \beta_T\mathrm{KL}_{\text{text}},
    \end{aligned}
\end{equation}
where $\mathcal{L}_{\text{inter}}$ inter-modal probabilistic NLL, averaged over the four image-text pairs, $\mathcal{L}_{\text{intra-I}}$ and $\mathcal{L}_{\text{intra-T}}$ are intra-modal symmetry losses between the image views and text inputs, $\mathrm{KL}_{\text{img}}$ and $\mathrm{KL}_{\text{text}}$ are variational information bottleneck (VIB) KL divergences for image and text embeddings, $\lambda_I$, $\lambda_T$, $\beta_I$, and $\beta_T$ are weight scalars.

This multi-view, multi-loss formulation provides richer supervision and produces probabilistic embeddings that are both semantically aligned and uncertainty-calibrated, ultimately improving cross-modal retrieval performance.

\subsection{Implementation Details}
We implement MedProbCLIP in \texttt{PyTorch}~\cite{paszke2019pytorch} following the model architecture introduced previously. For image encoding, we use a ViT\footnote{\texttt{ViT-Base/16}}~\cite{dosovitskiy2020vit}  backbonefrom \texttt{TIMM}~\cite{rw2019timm}, pretrained on ImageNet-21K~\cite{ridnik2021imagenet21k}. For text encoding, we adopt BioMedBERT\footnote{\texttt{microsoft/BiomedNLP-BiomedBERT-base-uncased-\\abstract-fulltext}}~\cite{pubmedbert} from \texttt{HuggingFace}~\cite{wolf2019huggingface}. Both encoders output deterministic embeddings that are fed into the distribution learning modules, each contains a shallow MLP, which produces the mean and log-variance vectors of a diagonal Gaussian distribution. Log-variances are clamped to $[-6, 6]$ for stability.

The model was trained using a Linux cluster with Nvidia A100 GPUs, with all random seeds fixed to 42 for reproducibility. We use the AdamW~\cite{loshchilov2017decoupledWD} optimizer with a base learning rate of $5e^{-5}$, weight decay $1e^{-4}$ , and gradient clipping ($\text{max-norm} =1.0$). Mixed-precision training is enabled through \texttt{torch.amp.GradScaler}. A cosine annealing learning-rate scheduler~\cite{loshchilov2016sgdr} is optionally used. All models are trained for 50 epochs with batch size 128.

\section{Experimental Result}

\subsection{Setup}
\subsubsection{Datasets}
We conduct our experiments on the MIMIC-CXR dataset~\cite{johnson2019mimic}, a large-scale publicly available corpus of chest radiographs and paired free-text radiology reports. The dataset contains 227,835 imaging studies from 64,588 patients, comprising 368,948 chest X-ray images along with their corresponding diagnostic reports. MIMIC-CXR also includes 14 clinical labels, 13 pathology categories plus a ``no finding" class, automatically derived from the reports using the NegBio and CheXpert NLP pipelines~\cite{peng2018negbio,irvin2019chexpert}.

A substantial portion of studies in MIMIC-CXR contain multiple imaging views, most commonly a PA view paired with a lateral view. Radiology reports likewise contain multiple semantically complementary sections, such as Findings and Impression. These intrinsic multi-view and multi-section structures closely align with our model design, which explicitly leverages multiple image and text inputs during training. When a study or report contains only a single view or section, we generate the missing counterpart through controlled data augmentation.

For preprocessing, all images are resized to $224\times224$ resolution and normalized using standard ImageNet statistics. Text reports are tokenized using the BioMedBERT tokenizer with a maximum sequence length of $256$ tokens, where truncation is applied to long reports and padding is used for shorter ones.

We follow the official dataset split: the training set is used for training, the validation set is used for checkpoint selection, and the held-out test set is used exclusively for final evaluation.

\vspace{-.5mm}
\subsubsection{Baseline Models}
\vspace{-.5mm}
We compare MedProbCLIP against three representative cross-modal retrieval baselines: CLIP, PCME++, and CXR-CLIP. To ensure a fair and controlled evaluation, we reimplement all baselines using the same visual and textual backbones as our method. All models are initialized from the same pretrained weights and fine-tuned on the MIMIC-CXR dataset under identical training settings.

\vspace{-4.mm}
\paragraph{CLIP}
We implement a deterministic CLIP-style dual encoder where image and text embeddings are projected into a shared latent space and trained using a bidirectional contrastive loss. This baseline captures strong deterministic image-text alignment without uncertainty modeling.

\vspace{-4.mm}
\paragraph{PCME++}
We reimplement the probabilistic cross-modal embedding model PCME++ using the same architecture as in our method but without multi-view modeling. PCME++ assigns a Gaussian distribution to each modality and optimizes the CSD. This baseline allows us to isolate the benefits of multi-view learning and architectural modifications introduced in MedProbCLIP.

\vspace{-4.mm}
\paragraph{CXR-CLIP}
CXR-CLIP extends CLIP to the clinical domain and incorporates multi-view learning by jointly encoding multiple chest X-ray views and multiple report sections. To ensure a fair comparison focused purely on retrieval, we omit all label-supervised components and retain only the multi-view contrastive image-text alignment. This provides a strong deterministic multi-view baseline aligned with our data format.

\vspace{-4mm}
\paragraph{Training}
All baseline models are trained using the same data splits, preprocessing steps, batch sizes, optimization settings, and evaluation procedures as MedProbCLIP, enabling a direct and fair comparison of retrieval performance.

\begin{table*}[!tb]
    \centering
    \setlength{\tabcolsep}{.925em}
    \begin{tabular}{c||c|c|c|c||c|c|c|c||c}
        \hline\hline
        \multirow{2}{*}{\textbf{Method}} & \multicolumn{4}{c||}{\textbf{i2t}} & \multicolumn{4}{c||}{\textbf{t2i}} & \multirow{2}{*}{\textbf{RSUM}}\\\cline{2-9}
        & \textbf{R@1} & \textbf{R@5} & \textbf{R@10} & \textbf{R@100} & \textbf{R@1} & \textbf{R@5} & \textbf{R@10} & \textbf{R@100} &  \\\hline\hline
        \textbf{Random} & $0.06$ & $0.28$ & $0.56$ & $5.62$ & $0.09$ & $0.46$ & $0.92$ & $9.20$ & $17.19$ \\\hline

        \textbf{CLIP} & $14.28$ & $33.73$ & $44.29$ & $80.94$ & $14.22$ & $34.23$ & $43.73$ & $80.89$ & $346.32$ \\\hline
        
        \textbf{PCME++} & $12.48$ & $31.53$ & $40.58$ & $77.68$ & $11.80$ & $32.32$ & $42.55$ & $78.58$ & $327.54$ \\\hline
        \textbf{CXR-CLIP} & $17.14$ & $41.65$ & $55.03$ & $89.94$ & $16.86$ & $41.26$ & $54.47$ & $90.39$ & $406.75$ \\\hline
        \textbf{MedProbCLIP} & $\textbf{21.02}$ & $\textbf{46.88}$ & $\textbf{58.91}$ & $\textbf{92.41}$ & $\textbf{19.96}$ & $\textbf{47.44}$ & $\textbf{59.42}$ & $\textbf{92.58}$ & $\textbf{438.62}$ \\\hline\hline
    \end{tabular}
    \vspace{-2.mm}
    \caption{Cross-modal retrieval performance on the MIMIC-CXR test set.}
    \label{tab:retrieval}
\end{table*}

\begin{table*}[!tb]
    \centering
    \setlength{\tabcolsep}{.25em}
    \small
    \begin{tabular}{c||c|c|c|c|c|c|c|c|c|c|c|c|c||c}
    \hline\hline
        \textbf{Method} & \textbf{EC} & \textbf{CA} & \textbf{LL} & \textbf{AO} & \textbf{ED} & \textbf{CO} & \textbf{PA} & \textbf{AT} & \textbf{PX} & \textbf{PE} & \textbf{PO} & \textbf{FR} & \textbf{SD} & \textbf{Mean} \\ \hline\hline

        \textbf{CLIP} & $0.6683$ & $0.7832$ & $0.4881$ & $\underline{0.7550}$ & $\textbf{0.8453}$ & $0.4792$ & $\underline{0.7236}$ & $0.3100$ & $0.6707$ & $\underline{0.6865}$ & $\underline{0.6364}$ & $\textbf{0.6364}$ & $\textbf{0.7925}$ & $0.6519$
 \\\hline
 
         \textbf{PCME++} & $0.6875$ & $\textbf{0.8061}$ & $0.1667$ & $0.4750$ & $0.7126$ & $0.5972$ & $0.4651$ & $\textbf{0.7150}$ & $0.4603$ & $0.5061$ & $0.4545$ & $0.5273$ & $0.5189$ & $0.5456$
 \\\hline

         \textbf{CXR-CLIP} & $\textbf{0.7837}$ & $0.7908$ & $0.7262$ & $\textbf{0.8167}$ & $\underline{0.8164}$ & $0.4722$ & $0.7144$ & $0.5000$ & $\underline{0.7336}$ & $0.5926$ & $0.4545$ & $0.4818$ & $0.7217$ & $\underline{0.6619}$
 \\\hline

         \textbf{MedProbCLIP} & $\underline{0.7163}$ & $\underline{0.7947}$ & $\textbf{0.7976}$ & $0.5767$ & $0.7986$ & $\textbf{0.7292}$ & $\textbf{0.7308}$ & $\underline{0.5650}$ & $\textbf{0.7802}$ & $\textbf{0.8661}$ & $\textbf{0.7273}$ & $0.4455$ & $\underline{0.7028}$ & $\textbf{0.7101}$\\\hline\hline
    \end{tabular}  
    \newline
    \flushleft
    {\footnotesize
    \vspace{-3mm}\hspace{1mm}EC: Enlarged Cardiomediastinum\hspace{2.5mm} CA: Cardiomegaly\hspace{2.5mm} LL: Lung Lesion\hspace{2.5mm} AO: Airspace Opacity\hspace{2.5mm} ED: Edema\hspace{2.5mm} CO: Consolidation\hspace{2.5mm} PA: Pneumonia\\
    \hspace{1mm}AT: Atelectasis\hspace{2.5mm} PX: Pneumothorax\hspace{2.5mm} PE: Pleural Effusion\hspace{2.5mm} PO: Pleural Other\hspace{2.5mm} FR: Fracture\hspace{2.5mm} SD: Support Devices}
    \vspace{-1.5mm}
    \caption{Zero-shot classification accuracy across 13 pathology categories on MIMIC-CXR.}
    \vspace{-1.mm}
    \label{tab:zero-shot}
\end{table*}

\subsubsection{Evaluation Metrics}
We assess the performance of all models across three complementary dimensions: retrieval quality, zero-shot generalization, and robustness.

\vspace{-4.mm}
\paragraph{Retrieval Performance}
We evaluate medical image-report retrieval using standard Recall@K metrics for $K\in\{1, 5, 10, 100\}$, computed separately for image-to-text (i2t) and text-to-image (t2i) retrieval. Following prior work, we also report RSUM, defined as the sum of all R@K scores across both retrieval directions, providing a single aggregated measure of overall retrieval quality.

\vspace{-4.mm}
\paragraph{Zero-Shot Classification}
To assess generalization to clinically meaningful semantic categories, we evaluate each model in a zero-shot classification setting on the 14 pathology labels provided by MIMIC-CXR. For each class $c$, we construct a text prompt of the form
``\texttt{a chest X-ray with $c$}" and compute the similarity between an image and all class prompts using the model’s image-text embedding space. The predicted label corresponds to the prompt with the highest similarity score. We report per-class classification accuracy, allowing a detailed assessment of how well the model captures each clinically relevant concept without any task-specific fine-tuning.

\vspace{-4.mm}
\paragraph{Robustness Evaluation}
We evaluate robustness from two perspectives.
First, we measure Selective Retrieval Performance, which assesses accuracy at varying coverage levels by allowing the model to abstain on low-confidence queries. This quantifies how reliably a model can identify high-certainty predictions. 
Second, we evaluate retrieval performance under image perturbations, including Gaussian blur, Gaussian noise, brightness/contrast jitter, and rotation. For each corruption type and severity, we report retrieval degradation relative to clean performance, enabling a systematic comparison of robustness across models.

\subsection{Result}
\vspace{-.5mm}
\subsubsection{Retrieval}

\begin{figure*}
    \centering
    \includegraphics[width=0.925\linewidth]{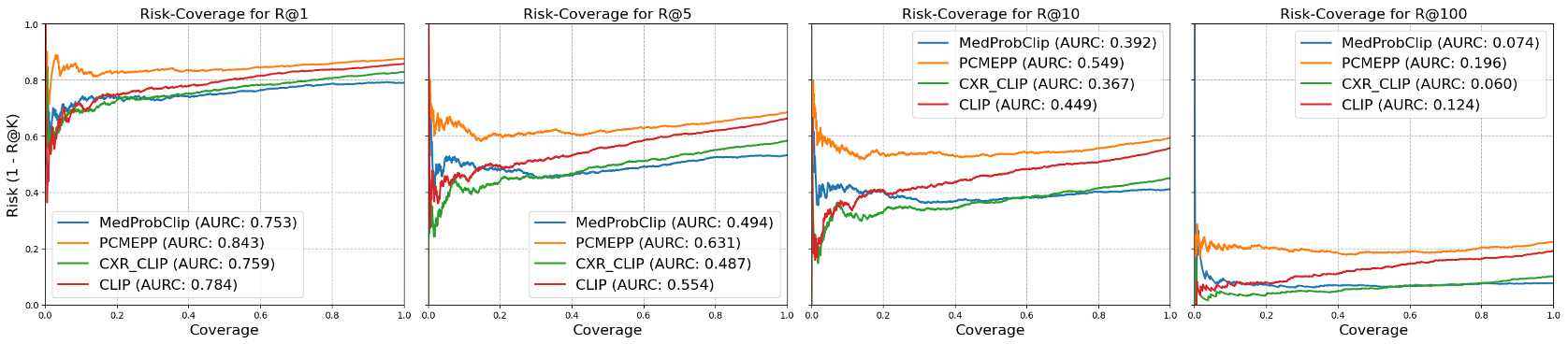}
    \vspace{-1.mm}
    \caption{Selective retrieval performance demonstrates MedProbCLIP’s superior calibration and potential for safer clinical deployment.}
    \vspace{-1.mm}
    \label{fig:selective_retrieval}
\end{figure*}
Table~\ref{tab:retrieval} reports medical image-report retrieval results on the MIMIC-CXR test set, measured using Recall@K for both i2t and t2i directions. MedProbCLIP achieves the best performance across all metrics, outperforming all deterministic and probabilistic baselines by a substantial margin.

Among the baselines, CXR-CLIP exhibits the strongest results, reflecting the benefit of multi-view modeling in clinical imaging. CLIP performs competitively but falls behind CXR-CLIP, while PCME++ performs slightly worse than CLIP despite its probabilistic formulation, likely due to its single-view training and less structured supervision.

MedProbCLIP delivers the highest retrieval accuracy in both directions. For i2t retrieval, it achieves R@1 = 21.02, improving over CXR-CLIP by 3.88 absolute points and over CLIP by 6.74 points. Similar gains are observed for t2i retrieval, where MedProbCLIP attains R@1 = 19.96, outperforming CXR-CLIP by 3.10 points. The advantages persist across all K values, demonstrating that MedProbCLIP consistently retrieves correct matches even at deeper rankerforms competitively and obtais.

Overall retrieval quality, summarized by RSUM, further highlights the improvement. MedProbCLIP achieves an RSUM of 438.62, surpassing CXR-CLIP by 31.87 points, CLIP by 92.30 points, and PCME++ by 111.08 points.


\vspace{-1.mm}
\subsubsection{Zero-Shot Classification}

Table~\ref{tab:zero-shot} summarizes zero-shot classification accuracy across pathology categories in MIMIC-CXR, using prompts of the form ``\texttt{a chest X-ray with \{\textit{class}}\}." MedProbCLIP achieves the highest mean zero-shot accuracy among all compared methods, demonstrating stronger semantic alignment between visual features and clinical concepts.

CXR-CLIP performs competitively and obtains the best accuracy on two categories and the second best on mean accuracy, reflecting its strong multi-view modeling. CLIP also achieves strong performance on certain high-level findings such as Support Devices and Fracture, but its overall mean accuracy remains lower due to weaker performance on more subtle pathologies (e.g., Atelectasis, Consolidation). PCME++ exhibits highly class-dependent behavior: it performs well on Cardiomegaly and Atelectasis but underperforms on classes requiring fine-grained image-text alignment (e.g., Lung Lesion).

MedProbCLIP delivers the best overall mean accuracy (0.7101), improving over CXR-CLIP by 4.82 points and CLIP by 5.82 points. It achieves the highest accuracy on 6 of the 13 classes, including Lung Lesion, Consolidation, Pneumonia, Pneumothorax, Pleural Effusion, and Pleural Other, where uncertainty modeling and multi-view supervision provide a greater advantage. These gains highlight that probabilistic embeddings enable the model to better capture subtle radiographic patterns and their textual descriptions, even in a zero-shot setting without task-specific fine-tuning.


\subsubsection{Selective Retrieval Performance}

To evaluate reliability under uncertainty, we measure Selective Retrieval Performance using Risk-Coverage curves (Figure~\ref{fig:selective_retrieval}). In this setting, the model may abstain from answering when confidence is low, allowing us to study how retrieval accuracy behaves as coverage, the fraction of answered queries-increases. Risk is defined as $1-R@K$, and overall reliability is summarized using the Area Under the Risk-Coverage Curve (AURC), where lower values indicate better calibration.

Across all recall levels (R@1, R@5, R@10, R@100), MedProbCLIP exhibits substantially better selective performance than the deterministic baselines (CLIP, CXR-CLIP) and the probabilistic PCME++ model. MedProbCLIP consistently yields the lowest or second-lowest AURC, demonstrating stronger uncertainty awareness and more trustworthy retrieval decisions.

A key advantage of MedProbCLIP is the stability of its risk curves: risk increases only gradually as coverage grows, indicating that the model maintains reliable predictions even when required to answer a larger share of queries. This behavior contrasts sharply with PCME++, whose risk rises rapidly with increased coverage, reflecting overconfident errors. CLIP and CXR-CLIP show better stability than PCME++, but their risk escalates more steeply than MedProbCLIP, particularly at low-coverage regions where calibrated confidence matters most.


\subsubsection{Robustness to Image Perturbations}

\begin{figure*}[!tb]
    \centering    
     
     \begin{subfigure}[b]{0.9\textwidth}
         \centering
         \includegraphics[width=\textwidth]{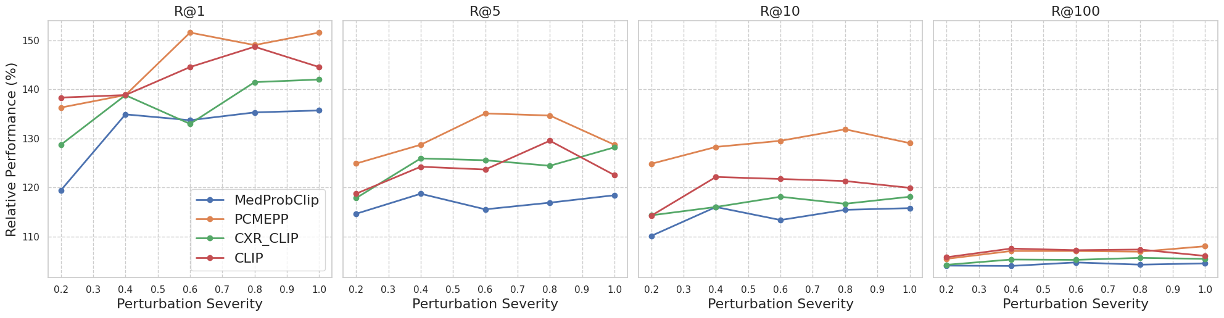}
         \caption{Relative performance degradation under Gaussian blur across severity levels.}
         \label{fig:blur}
     \end{subfigure}

     \begin{subfigure}[b]{0.9\textwidth}
         \centering
         \includegraphics[width=\textwidth]{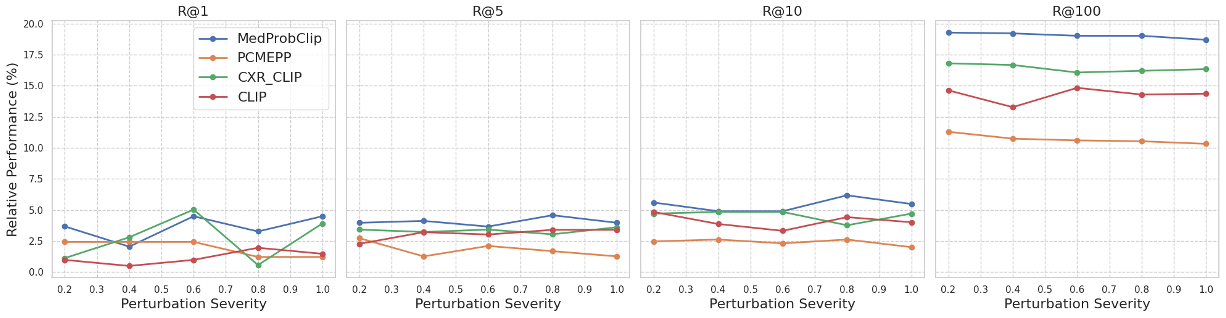}
         \caption{Relative performance degradation under Gaussian noise across severity levels.}
         \label{fig:noise}
     \end{subfigure}

     \begin{subfigure}[b]{0.9\textwidth}
         \centering
         \includegraphics[width=\textwidth]{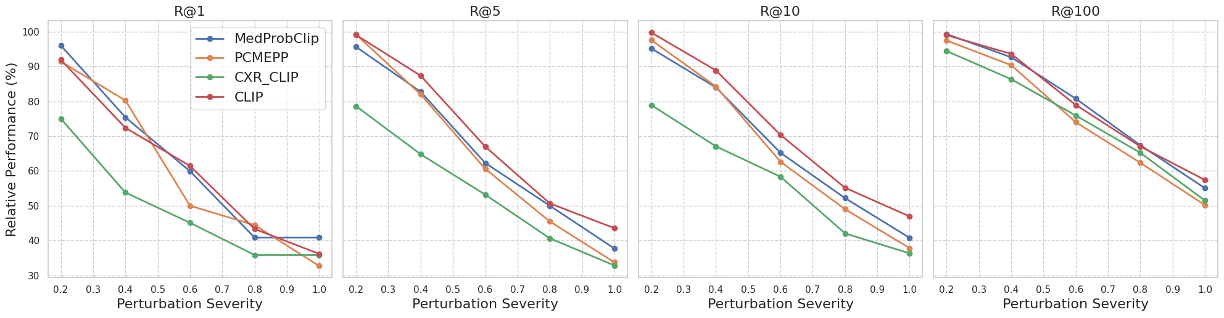}
         \caption{Relative performance degradation under brightness/contrast shifts.}
         \label{fig:brightness}
     \end{subfigure}

     \begin{subfigure}[b]{0.9\textwidth}
         \centering
         \includegraphics[width=\textwidth]{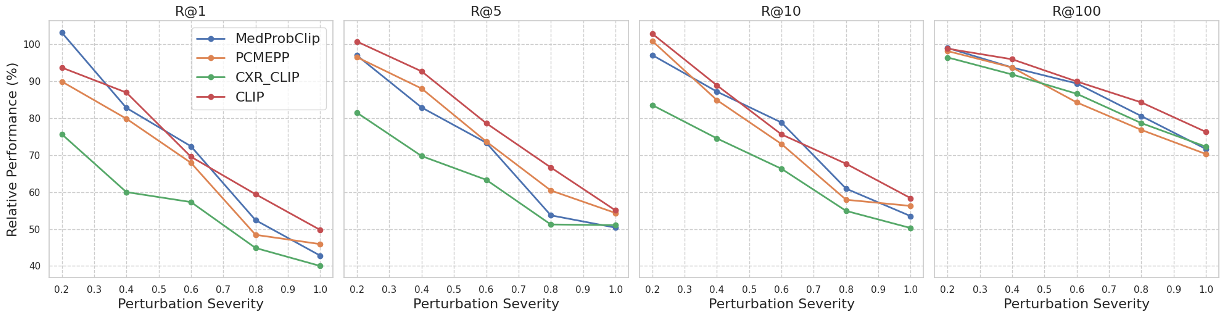}
         \caption{Relative performance degradation under rotation perturbations.}
         \label{fig:rotation}
     \end{subfigure}

     \vspace{-1mm}
     \caption{Robustness of retrieval performance under four categories of image perturbations.
     }
     \vspace{-4mm}
     \label{fig:robustness}
\end{figure*}

We further examine robustness by evaluating retrieval performance under four clinically relevant perturbations: Gaussian blur, Gaussian noise, brightness/contrast shifts, and rotation, each applied at multiple severity levels. Figure~\ref{fig:robustness} shows the relative Recall@K performance compared to the clean baseline.

\vspace{-4.5mm}
\paragraph{Gaussian Blur}
Gaussian blur removes high-frequency structure that is essential for radiologic interpretation. Although some baselines (e.g., PCME++ or CLIP) exhibit spuriously higher relative performance at certain blur levels, this is not a sign of robustness. In fact, it reflects instability in retrieval rankings rather than genuine invariance. Blur strongly disrupts all deterministic baselines, producing large fluctuations across severity levels. In contrast, MedProbCLIP exhibits the smoothest and most stable degradation, avoiding the large swings seen in PCME++ and CLIP. This indicates that MedProbCLIP’s uncertainty-aware embeddings are less sensitive to loss of local detail.

\vspace{-4.5mm}
\paragraph{Gaussian Noise}
Gaussian noise severely corrupts edge information and pixel-level gradients, making it the most destructive perturbation. All models degraded substantially, even at mild noise levels. Despite this difficulty, MedProbCLIP consistently achieves the highest relative performance across recall thresholds. Its degradation curve is also the most stable and least erratic, whereas PCME++ and CLIP show unpredictable fluctuations and sharp drops. This demonstrates that probabilistic modeling provides greater resilience to noise-induced pixel distortions than deterministic or purely distance-based embeddings.

\vspace{-4.5mm}
\paragraph{Brightness Contrast Shifts}
These perturbations mimic variations in scanner calibration and acquisition conditions. All models show monotonic degradation as severity increases. Although performance drops sharply at extreme severities, MedProbCLIP generally ranks among the most stable methods across recall levels, demonstrating resilience to global photometric distortions and improved tolerance to intensity variations relative to PCME++ and CXR-CLIP.

\vspace{-4.5mm}
\paragraph{Rotation}
Rotational perturbations disrupt anatomical orientation and spatial alignment. All models degrade as rotation increases. MedProbCLIP shows slower degradation than PCME++ and CXR-CLIP across most settings, reflecting improved spatial robustness. While CLIP benefits from large-scale natural-image pretraining and performs competitively at extreme rotations, MedProbCLIP achieves more consistent robustness across the entire severity spectrum, especially at moderate rotations typical in clinical variation.

\section{Discussion and Conclusion}
Medical image-text alignment inherently involves ambiguity, multi-view inconsistencies, and imperfect supervision, making deterministic embedding spaces prone to overconfident and brittle representations. MedProbCLIP addresses these challenges through probabilistic contrastive learning, which allows the model to represent not only what features it extracts but also how certain it is about them.

A key advantage of MedProbCLIP is its ability to allocate uncertainty where it is needed. When radiographs contain subtle findings, heterogeneous views, or partial evidence, the model naturally increases variance, preventing overconfident mistakes and improving calibration. This produces smoother decision boundaries, more reliable similarity comparisons, and more stable behavior as coverage increases in selective retrieval. Probabilistic modeling also acts as a form of structured regularization, reducing sensitivity to spurious correlations and mitigating abrupt performance drops under moderate perturbations such as blur, photometric shifts, or rotations. In multi-view settings, common in clinical imaging, the probabilistic formulation helps consolidate complementary cues while preventing one view from dominating the representation.

MedProbCLIP is not without limitations. When the supervision signal is clean and unambiguous, deterministic models may achieve similar performance without the overhead of variance estimation. Under extreme corruptions such as heavy Gaussian noise, uncertainty modeling cannot compensate for the collapse of visual signal, although it may still degrade more smoothly. Moreover, probabilistic models rely on careful variance parameterization and KL regularization; poor tuning can lead to under- or over-dispersed embeddings.

Despite these constraints, the advantages of probabilistic contrastive learning are substantial in clinical retrieval scenarios where uncertainty is intrinsic rather than incidental. By explicitly modeling ambiguity, MedProbCLIP yields more calibrated, more robust, and more semantically aligned cross-modal representations. Future work may explore richer uncertainty structures, adaptive calibration, or integration with downstream clinical decision-support systems, where principled uncertainty remains essential.

\section*{Acknowledgments}
This material is based upon work supported by the National Science Foundation under Award No. 2334243. Any opinions, findings and conclusions or recommendations expressed in this material are those of the author(s) and do not necessarily reflect the views of the National Science Foundation.

\vspace{1.5mm}
\noindent The computational portion of this research utilized the GPU cluster provided by the High Performance Computing Research Center at Texas A\&M University-San Antonio. We gratefully acknowledge the center's support in facilitating access to these resources. 
{
    \small
    \bibliographystyle{IEEEtran}
    \bibliography{main}
}

\end{document}